\documentclass[journal]{IEEEtran}
\usepackage{threeparttable}

\usepackage{cite}

\usepackage{graphicx}

\usepackage[cmex10]{amsmath}

\usepackage{algpseudocode}
\usepackage[caption=false,font=footnotesize]{subfig}

\usepackage{amssymb}

\allowdisplaybreaks

\begin{document}
%
\title{A Probabilistic Framework for Representing Dialog Systems and Entropy-Based Dialog Management through Dynamic Stochastic State Evolution}
%
%
%
\author{Ji~Wu,~\IEEEmembership{Senior Member,~IEEE,}
		Miao~Li,
        Chin-Hui Lee,~\IEEEmembership{Fellow,~IEEE} 
\thanks{This work was done while the first author was visiting School of Electrical and Computer Engineering, Georgia Institute of Technology in Atlanta, Georgia, USA, in 2013-2015.}
\thanks{Ji Wu and Miao Li are with the Multimedia Signal and Intelligent Information Processing Laboratory, Department of Electronic Engineering, Tsinghua University, Beijing 100084, China (e-mail: wuji\_ee@tsinghua.edu.cn, limiaogg@126.com).}
\thanks{ Chin-hui Lee is with the School of Electrical and Computer Engineering, Georgia Institute of Technology, Atlanta, GA. 30332-0250, USA (e-mail: chinhui.lee@ece.gatech.edu).}
}
\maketitle

\begin{abstract}
In this paper, we present a probabilistic framework for goal-driven spoken dialog systems. A new dynamic stochastic state (DS-state) is then defined to characterize the goal set of a dialog state at different stages of the dialog process. Furthermore, an entropy minimization dialog management(EMDM) strategy is also proposed to combine with the DS-states to facilitate a robust and efficient solution in reaching a user's goals. A Song-On-Demand task, with a total of 38117 songs and 12 attributes corresponding to each song, is used to test the performance of the proposed approach. In an ideal simulation, assuming no errors, the EMDM strategy is the most efficient goal-seeking method among all tested approaches, returning the correct song within 3.3 dialog turns on average. Furthermore, in a practical scenario, with top five candidates to handle the unavoidable automatic speech recognition (ASR) and natural language understanding (NLU) errors, the results show that only 61.7\% of the dialog goals can be successfully obtained in 6.23 dialog turns on average when random questions are asked by the system, whereas if the proposed DS-states are updated with the top 5 candidates from the SLU output using the proposed EMDM strategy executed at every DS-state, then a 86.7\% dialog success rate can be accomplished effectively within 5.17 dialog turns on average. We also demonstrate that  entropy-based DM strategies are more efficient than non-entropy based DM. Moreover, using the goal set distributions in EMDM, the results are better than those without them, such as in sate-of-the-art database summary DM.
\end{abstract}


%
\IEEEpeerreviewmaketitle

\section{Introduction}
\label{SecIntro}
%
%
%
%

\IEEEPARstart{S}{poken} dialog systems enable a human user to acquire information and services by interacting with a computer agent using spoken languages \cite{jokinen2009spoken, zue2000conversational}. Many such systems have been implemented to provide a variety of services, such as call routing \cite{lee2000natural}, flight booking \cite{seneff2000dialogue}, weather forecasting \cite{goddeau1994galaxy} and restaurants recommendation \cite{weng2007chat}. In the era of today's mobile internet, with the rapid technological advances in automatic speech recognition (ASR) \cite{rabiner1993fundamentals,povey2011kaldi,hinton2012deep} and natural language understanding (NLU) \cite{zue2000conversational, rosenfield2000two, huang2001spoken, mesnil2013investigation}, voice assistant applications, such as Apple's Siri \cite{bellegarda2014spoken}, Google's Voice Actions\footnote{http://www.google.com/mobile/voice-actions}, and iFlyTek Lingxi\footnote{http://www.iflytek.com/en/mobile/lingxi.html}, have begun to change individuals' daily life. The systems mentioned above are all task-oriented, focusing on solving practical problems for users in various domains.

Broadly speaking, a spoken dialog system consists of three major components. First, an \emph{input module} handles the input utterances from a user and attempts to extract the user intentions. Then, a \emph{control module} manages a dialog session and decides what actions the system should take to reach the user's goal \cite{scheffler1999simulation}. Finally, an \emph{output module} executes the system actions for the user. This paper primarily focuses on the control module.

A dialog is an interactive process between a user and a computer agent. In such a process,  three basic concepts apply, namely: \emph{state}, \emph{action} and \emph{policy} \cite{jokinen2009spoken}. A state is commonly used to describe the current circumstances of a dialog process \cite{scheffler1999simulation}. Typically, a dialog state can be factored into a number of simple discrete components that are extracted from distinct sources, such as a \emph{dialog goal}, a set of \emph{user's inputs} and a \emph{dialog history} \cite{young2010hidden}. An action often refers what the system can perform to interact with the user, e.g., a query to solicit further information from the user \cite{jokinen2009spoken}. Finally, a policy usually characterizes the strategy used by a system to determine which action to take in the current system state \cite{williams2007partially}.

To achieve the dialog goal successfully and efficiently, \emph{dialog management} (DM) performed by the control module is regarded as a core issue related to the design of a spoken dialog system. First, the DM module must maintain the states of the dialog process, considering both the dialog history and the user's inputs. Second, the DM module must choose a system action based on the current state in accordance with a certain policy. Conventional methods \cite{pieraccini2005we,pellom2001university,hanna2007promoting} generally handle the dialog processes in a deterministic way. They assume that the system must be in a certain predefined state, and a system action is chosen according to some manually specified rules. Many graph-based \cite{pieraccini2005we} and frame-based systems \cite{pellom2001university,hanna2007promoting} are typical examples of these so-called rule-based techniques. They are often effective and easy to implement for simple tasks. However, it can become extremely difficult to design rules for complex scenarios.

Recently, several approaches have been proposed that attempt to determine the optimal actions through statistical models \cite{Schatzmann2006Survey} instead of manually specified rules. These models are generally based on the Markov decision process (MDP) \cite{levin1998using,levin2000stochastic,young2000probabilistic}, and the DM strategies that are used to select actions based on the current state are optimized through reinforcement learning \cite{barto1998reinforcement}. To address ASR and NLU errors, an improved framework called the partially observable Markov decision process (POMDP) \cite{roy2000spoken} framework has been proposed. Rather than maintaining one hypothesis for the state, the POMDP framework maintains a probability distribution over all possible dialog states \cite{williams2007scaling}. Several approaches have been developed to track hidden dialog states \cite{williams2007partially,henderson2008mixture,young2010hidden,henderson2013deep}. Scalability is assumed to be a major difficulty of the POMDP framework. The number of potential states grows exponentially during a dialog process, causing the computational complexity of exact POMDP optimization to increase astronomically \cite{williams2007scaling}. Because of the difficulties of collecting sufficient real dialog data, the training of the dialog policies typically relies on user simulators \cite{gasic2014gaussian}. Because the behaviour of the simulated users is not perfectly realistic, the resulting policies are likely to significantly under-performing when compared with the policies trained on real interactive data \cite{crook2014real}.

Other researchers have attempted to exploit data mining techniques to summarize the back-end database and generate specific queries to address to users in a cooperative manner \cite{polifroni2006learning,polifroni2008intensional}. In such an approach, the DM module scans the database in every turn of the dialog process. It filters the database based on the user's intentions and preferences, and the resulting items are used to determine the content of the summary and the system actions. These resulting queries tend to ask users about the attributes with the highest uncertainty \cite{polifroni2006learning}, as the answers to these queries should allow for a maximum reduction of the search space. It was demonstrated  that the dialog process will be more efficient if the dialog system can generate questions that are suitable for the user based on the intentional summaries. One shortcoming of this database summary dialog management (DSDM) approach is that it implicitly assumes that all attributes are uniformly distributed. However, this assumption prevents the integration of a realistic prior distribution of the database entries into DM in which those attributes with higher entropies \cite{polifroni2006learning} could be taken into account.

In this paper, we first present a probabilistic framework for representing spoken dialog systems. A dynamic stochastic state (DS-state) is then defined based on the distribution over the goal set of a dialog. Such a state is capable of characterizing the overall situation of a dialog process more accurately and efficiently when compared with the state definitions used in state-of-the-art dialog systems (e.g., \cite{young2000probabilistic,roy2000spoken}). Leveraging on the probabilistic framework and the new DS-state, an entropy minimization dialog management (EMDM) strategy is finally proposed. It will be shown in the paper that this entropy-based approach is the best strategy for achieving the dialog goals effectively with the highest dialog success rate and efficiently with the least number of dialog turns in both ideal and practical systems.

We test the proposed approach on a Song-On-Demand (SoD) task with an access to a total of  38117 songs with a set of 12 attributes associate with each song. In ideal conditions without ASR and NLU errors, 8.3 dialog turns are required on average to return the correct song if random questions are asked by the dialog system. In comparison, the proposed EMDM strategy proves to be very efficient, requiring only 3.3 dialogue turns on average to identify a song. Furthermore, if the prior distribution over the goal set is considered, the proposed approach can provide a much better performance than other strategies. In a real testing scenario, there will be unavoidable ASR and NLU errors in the outputs of the SLU module. Such errors might cause the dialog task to fail or to require more turns of interaction resulting in only a small dialog success rate of 61.7\% when random questions are asked by the dialog system, achieving the dialog goals within 6.23 turns on average. To improve the system performance, a set of top-5 candidates from the SLU results is used to update the DS-states, and the EMDM strategy is then utilized to generate the next question. It is found that 86.7\% of the dialogs can be successfully completed within 5.17 dialogue turns on average. This represents a significant improvement over the random strategy in success rate and with fewer dialog turns.

The remainder of the paper is organized as follows. Section \ref{SecProbFrame} details the probabilistic framework. Section \ref{SecData} describes the task and the corresponding database used for concept illustration and experimental evaluations. Then, the proposed dynamic stochastic state is first introduced in Section \ref{SecSt}. Next the proposed entropy minimization dialog management strategy is presented in Section \ref{SecEMDM}. The evolution of the DS-states based on the multiple candidates outputs of the SLU module with potential ASR and SLU errors is illustrated in Section \ref{SecDSS}. Subsequently, Section \ref{SecExperiment} describes the experimental setup and presents an analysis of the results. Finally, we conclude our findings in Section \ref{SecConclusion}.




\section{Probabilistic Dialog Representation}
\label{SecProbFrame}
Typically, there is a structured database $D$ on the back-end of any goal-driven information access system. The entries and their corresponding information on $D$ represent the potential dialog goals, denoted by ${\bf G}=\left\{g_i|i=1,2,...,I\right\}$, that are sought by a user of the dialog system. $I$ denotes the number of entities in ${\bf G}$. Each database entry $g_i$ is often associated with a common set of attributes ${\bf A}=\left\{a_k|k=1,2,...,K\right\}$, where $K$ is the number of attributes. In a dialog process, to obtain the specific information and help to reach the user's goal quickly, the system presents a sequence of questions, which can be denoted by ${\bf Q}=\left\{q^{(j)}|j=1,2,...,J\right\}$. Then, the user provides a sequence of responses, denoted by ${\bf R}=\left\{r^{(j)}|j=1,2,...,J\right\}$. Each $r^{(j)}$ is the user response to the system question $q^{(j)}$. The pair $(q^{(j)},r^{(j)})$ forms an interaction, or a dialog turn of a dialog process. To begin a dialog, the system is usually in the initial state, $S^{(0)}$. After each turn of interaction, the dialog process evolves with a sequence of new states as follows: $S^{(1)}, S^{(2)},...,S^{(j)},...S^{(J)}$. Thus, the state sequence can be denoted by ${\bf S}=\mathbb{S}_0^J=\left\{S^{(j)}|j=0,1,...,J\right\}$. If $J=1$, then this dialog process consists of a single interaction, and the dialog system is a one-shot system. If $J>1$, then it is a multiple-interaction dialog system, and the probability of the entire dialog process is represented by:

\begin{equation}
\begin{aligned}
& P({\bf S},{\bf Q},{\bf R},D)=P(\mathbb{S}_0^J, \mathbb{H}_1^J, D)
\end{aligned}
\label{PbFmEQ1}
\end{equation}
where $\mathbb{H}_1^j = \{(q^{(1)},r^{(1)}),(q^{(2)},r^{(2)}),...,(q^{(j)},r^{(j)})\}$ denotes a dialog interaction history represented by a sequence of pairs up to state $S^{(j)}$. The primary objective in managing the dialog process is to generate a system question $q^{(j)}$ at the $j^{th}$ dialog interaction based on the back-end database $D$ and the past interaction history, $\mathbb{H}_1^{(j-1)}$. Thus, dialog management represents a strategy for accomplishing the user's intended goal while optimizing a given set of performance metrics, such as maximizing the dialog success rate and minimizing the required number of dialog interactions, $J$.

In the initial state, $S^{(0)}$, each potential goal, $g_i(i=1,2,....I)$, in the back-end database $D$, has a prior probability: $P_i^{(0)}=P(g_i|D)$. In the $(j-1)^{th}$ turn of the interaction, based on the immediate past state, $S^{(j-1)}$, the dialog system can generate the next question, $q^{(j)}$, with a probability of $P(q^{(j)}|S^{(j-1)},D)$. Depending on an understanding of the corresponding user input $r^{(j)}$, the interaction history,  $\mathbb{H}_1^{j-1}$, the DS-state evolution, $\mathbb{S}_0^{j-1}=\{S^{(0)},...,S^{(j-1)}\}$, and the back-end database $D$, the dialog process will reach the state $S^{(j)}$ with a probability of $P(S^{(j)}|(q^{(l)},r^{(l)}){_{l=1}^j},D)$. In the new state $S^{(j)}$, each potential goal, $g_i(i=1,2,....I)$, has a conditional probability, $P_i^{(j)}=P(g_i|S^{(j)},D)$. In view of the proposed probabilistic framework, we can now express the probability of the current dialog situation conditioned on the past as a product of three probabilities shown below:
\begin{equation}
\begin{aligned}
& P(q^{(j)}, r^{(j)},S^{(j)}|\mathbb{S}_0^{(j-1)},\mathbb{H}_1^{j-1},D) \\
& = P(q^{(j)}|S^{(j-1)},D) * P(r^{(j)}|q^{(j)},D) \\
&\hspace{1.1em} * P(S^{(j)}|S^{(j-1)},\mathbb{H}_1^j,D)
\end{aligned}
\label{PbFmEQ2}
\end{equation}
where the last probability term in the RHS of Eq. (\ref{PbFmEQ2}),  $P(S^{(j)}|S^{(j-1)},\mathbb{H}_1^j,D)$, realizes the first function of the DM model to maintain the DS-state evolution of the dialog process which will be illustrated in detail in Sections \ref{SecSt} and \ref{SecDSS} later. Meanwhile, the probability, $P(q^{(j)}|S^{(j-1)},D)$, in the first term of the RHS in Eq. (\ref{PbFmEQ2}) characterizes the second function of the DM model for determining the next question based on the past state. In addition to the two components of the DM model, Eq. (\ref{PbFmEQ2}) also contains a probability, $P(r^{(j)}|q^{(j)},D)$, the second term in the RHS, which can be regarded as the response model. For example, some questions are easy for users to answer, whereas others are not. In this study, we do not take this model into consideration. We simply assume that the users are collaborative and knowledgeable to always provide the correct answers to system questions.

For the initial state $S^{(0)}$, the entropy on the goal set ${\bf G}$ is:
\begin{equation}
H^{(0)}=-\sum\limits_{i=1}^I P_i^{(0)} {\rm log} P_i^{(0)}.
\label{PbFmEQ3}
\end{equation}
After the $j^{th}$ turn, the entropy at $S^{(j)}$ is represented by
\begin{equation}
H^{(j)}=-\sum\limits_{i=1}^I P_i^{(j)} {\rm log} P_i^{(j)}
\label{PbFmEQ4}
\end{equation}
and we can expect that:
\begin{equation}
H^{(0)}>H^{(1)}>...>H^{(j)}>...>H^{(J)} \geq 0
\label{PbFmEQ5}
\end{equation}
i.e., the system progresses in a direction of reducing the goal uncertainty, or entropy,
which is at a minimum when the goal is correctly and successfully reached. Therefore, a key objective of designing a good dialog system is equivalent to finding a strategy to quickly minimize the entropy.

\section{Task and Database Description}
\label{SecData}
Different from the conventional presentation flow, we describe the task and database in advance to make it easy to illustrate concepts of of proposed DS-state and EMDM strategy. In this study, a Song-On-Demand (SoD) task is used to test the proposed probabilistic framework. The back-end database consists of 38117 songs in total and each song is associated with a set of 12 attributes listed in Table \ref{12ATTRs}. These data were primarily collected from the internet and other sources. In the SoD task, a user attempts to find a song based on certain attributes of that song, and the system returns the correct song based on the information provided by the user. Although this is a simple task, it faces similar challenges in common with other spoken dialog tasks. 

\begin{figure}[h]
\centering
\subfloat[]{\includegraphics[width=2in, height=0.45in]{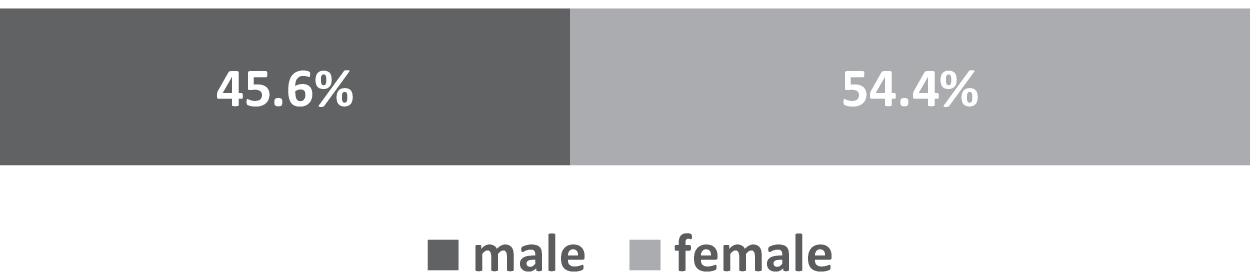}%
\label{GENDER}}
 \hfil
\subfloat[]{\includegraphics[width=2.5in, height=0.75in]{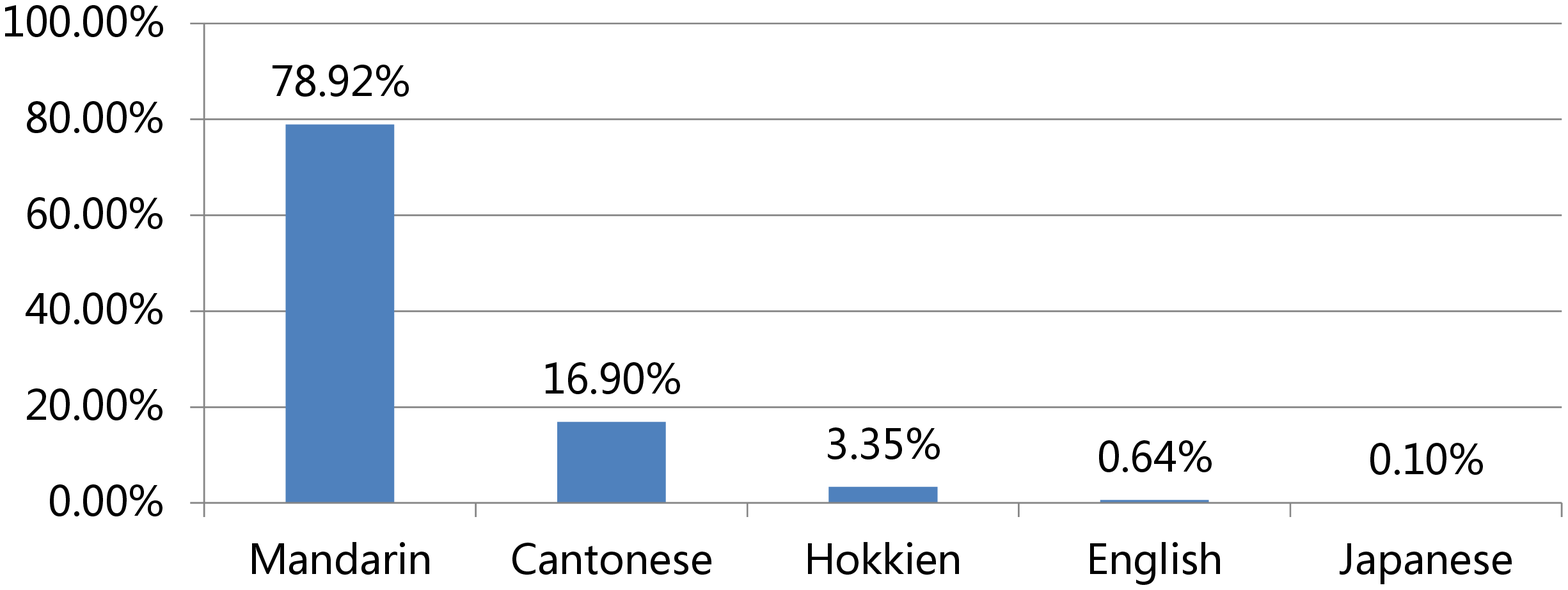}%
\label{LANGUAGE}}
 \hfil
\subfloat[]{\includegraphics[width=2.5in, height=0.75in]{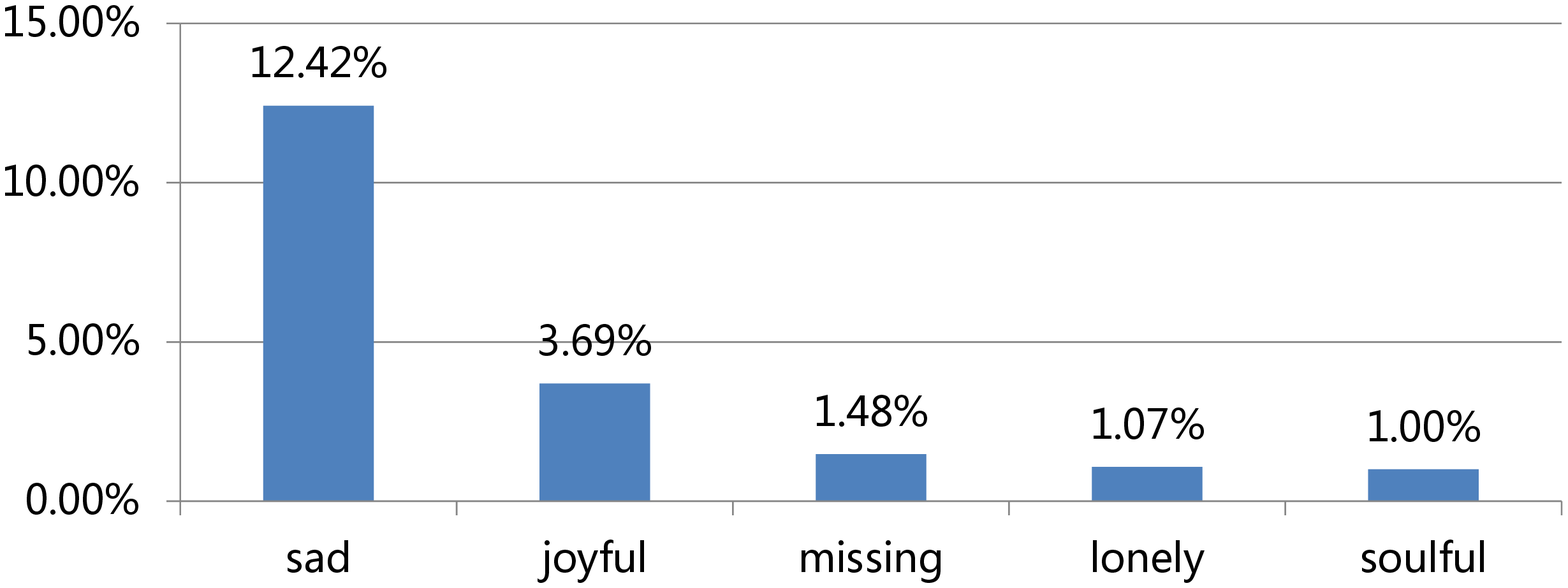}%
\label{EMOTION}}
\caption{The distributions of several attributes in the music database; (a) distribution of the \emph{Gender} attribute (ID 2), (b) partial distribution of the top 5 categories of the \emph{Language} attribute (ID 6), (c) partial distribution of the top 5 categories of the \emph{Emotion} attribute (ID 12).}
\label{DMDATA_Fig}
\end{figure}

 \begin{table}[h]
\renewcommand{\arraystretch}{1.3}
\caption{The 12 attributes of a song}
\label{12ATTRs}
\centering
\resizebox{0.48\textwidth}{!}{
\begin{tabular}{@{}|c||c|c|c|}
\hline
\textbf{ID} & \textbf{Attributes} & \textbf{Description} & \textbf{Value Numbers} \\
\hline
\hline
\emph{1} & \emph{Singe}r & The name of the singer & 3021\\
\hline
\emph{2} & \emph{Gender} & The gender of the singer & 2\\
\hline
\emph{3} & \emph{Region} & The region of the singer& 19\\
\hline
\emph{4} & \emph{Albu}m & {The album on which the song appears} & 10322\\
\hline
\emph{5} & \emph{Company} & The publisher of the song & 1193\\
\hline
\emph{6} & \emph{Language} & The language of the song & 10 \\
\hline
\emph{7} & \emph{Lyricist} & The lyricist of the song & 5603 \\
\hline
\emph{8} &\emph{ Composer} & The composer of the song & 5642 \\
\hline
\emph{9} & \emph{Live} &Live version or not & 2\\
\hline
\emph{10} &\emph{ Time} & The release date of the song & 413\\
\hline
\emph{11} & \emph{Style} & The style of the song & 346 \\
\hline
\emph{12} & \emph{Emotion} & The emotion of the song & 59 \\
\hline
\end{tabular}
}
\end{table}
To characterize the music database, we take advantage of the fact that the distributions of these 12 attributes are different, as are their corresponding entropies. For example, the numbers of possible values for each of the 12 attributes are shown in the rightmost column of Table \ref{12ATTRs}, and the collection of 38117 songs correspond to 10322 albums and 3020 singers. Some statistical information about these key attributes can be found in Fig. \ref{DMDATA_Fig}. For example, there are more songs from female singers (at 54.4\%) than from male singers shown in Fig. \ref{DMDATA_Fig}(a). Most of the songs are in Chinese, including Mandarin, Cantonese and Hokkien shown in Fig. \ref{DMDATA_Fig}(b). Moreover the ''\emph{sad}'' tag ranks first in the \emph{emotion} attribute as shown in Fig. \ref{DMDATA_Fig}(c).

It is a difficult task to collect all relevant values of these attributes for so many songs. Some attribute information is missing for some songs. The three attributes of \emph{Singer}, \emph{Album} and \emph{Time} are the best-presented attributes. Almost all songs have values for these three attributes. In contrast, the \emph{Style}, \emph{Composer} and \emph{Emotion} categories are the top three attributes whose representations are missing among the songs in the database, with a missing rate of 53\%, 50\% and 20\%, respectively. Because of the missing values of some attributes, the final goal set may include more than one song, all of which satisfy the requirements of the users.

\section{Dynamic Stochastic State in Dialog Systems}
\label{SecSt}
\begin{figure}[h]
\centering
\includegraphics[width=3.5in, height=1.8in]{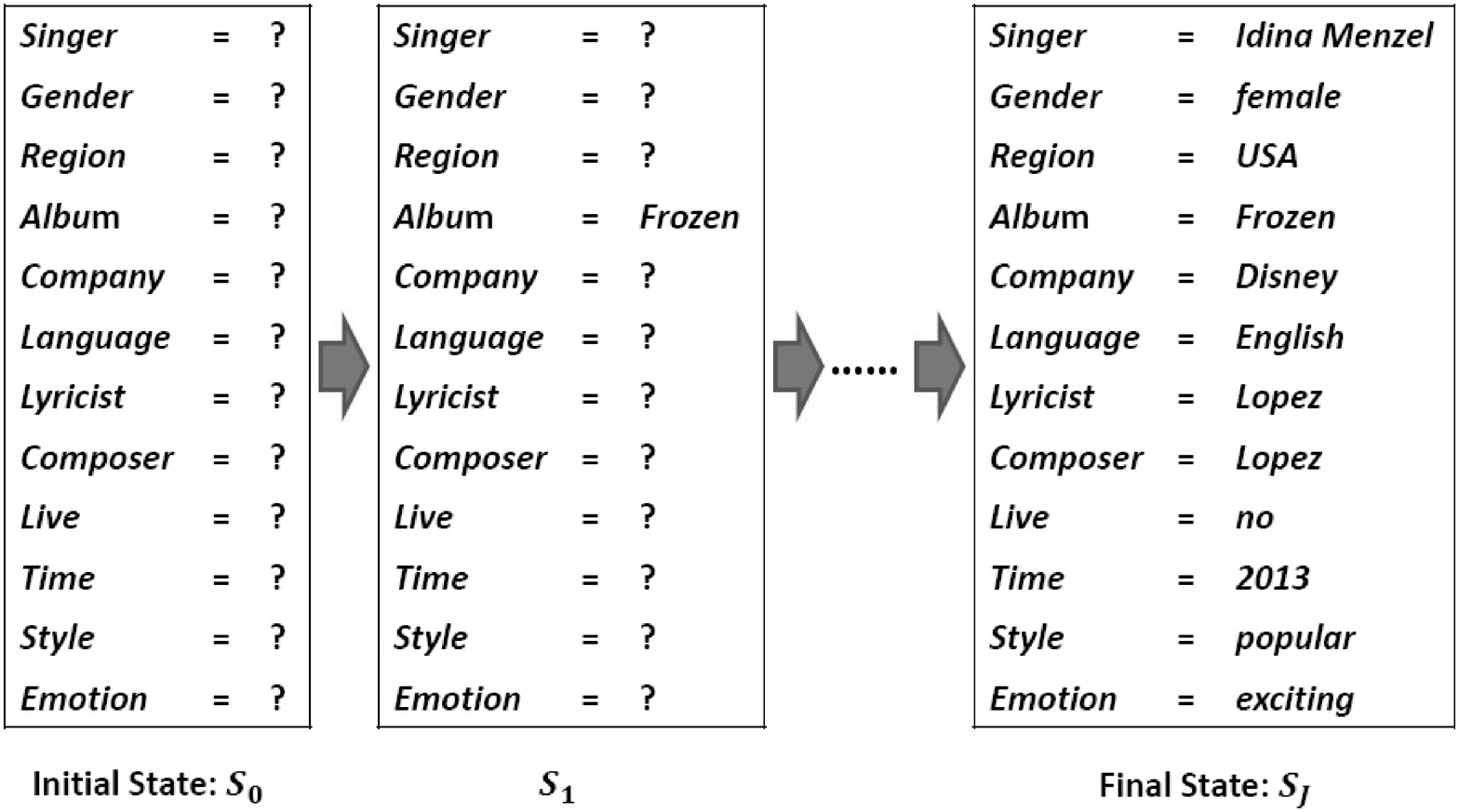}
\caption{An example of state sequences based on the dialogue state definition in the MDP/POMDP framework}
\label{FigTradSt}
\end{figure}

The state definition plays an important role in characterizing a dialog system. It not only describes the system's situations but also partially determines the next actions that the system may execute \cite{levin1998using}. In most current studies, such as those based on the MDP/POMDP framework, each state is represented by the values of all relevant internal variables \cite{levin2000stochastic}. Taking our SoD task as an example, the state description should include at least 12 variables that summarize all attributes of a song. In each interaction turn, after receiving additional information from the user's inputs, the system reaches its next state. Fig. \ref{FigTradSt} presents a state sequence example used to reach the goal, "Let It Go", a very popular theme song of Disney's "Frozen".

In the MDP framework, because the state description can be factored into a number of simple discrete components \cite{young2010hidden}, the states must be discrete and defined in advance in accordance with the task. For most practical dialog systems, the states are complex, and hence, the full state space would be intractably large \cite{young2010hidden}. To address the ASR and NLU errors, a "belief state" is introduced to indicate the probability distribution over all states at a specific time \cite{williams2007partially} in the POMDP framework. Therefore, scalability is a great challenge in the POMDP approach because of the curse of dimensionality and an increasing complexity in a long dialog history \cite{pineau2003point}.

In this paper, we attempt to depict dialog states in a different way. We use a distribution over the goal set to represent a state. This distribution is the result of the entire dialog interaction history and represents all the information available in that state. Moreover, the next system action jointly depends on the distribution and the DM strategy. Thus, the initial state $S^{(0)}$ can be represented by the prior distribution of the original goal set, and the uncertainty continues to decrease throughout the dialog process, until in the final state $S^{(J)}$, the dialog process reaches the goal with a minimal uncertainty. For a particular state in the MDP/POMDP framework, each attribute must be assigned a value, whereas for the states we define here, the value of each attribute is also a random variable which can be calculated from the distribution over the goal set. We argue that the states defined in this way can represent the circumstances of a dialog process more accurately and efficiently. Furthermore, when this definition is used, the dialog states can no longer be defined in advance. Instead, they will be generated step-by-step during the dialog process. We term the dialog state here a Dynamic Stochastic State (DS-state). In the remainder of this paper, we use the symbol sequence, $S^{(1)}, S^{(2)},...,S^{(j)},...S^{(J)}$, to represent the DS-state sequence and its evolution through the dialog process will be described in detail later in Section \ref{SecDSS}.

\section{Entropy Minimization Dialog Management}
\label{SecEMDM}
The attributes are the primary source of information that can help to narrow down the candidate list based on a particular user entry. Therefore, the questions presented by the dialog systems should make a good use of these attributes. To simplify the system strategy, we assume that each question is related to only one of the attributes. In this case, if a user works collaboratively with the system, the user's answers should be related to the same attributes as do the corresponding questions. The prior probability of a potential goal $g_i$ is $P_i^{(0)}=P(g_i)$, and the conditional probability of $g_i$ for DS-state $S^{(j)}$ is $P_i^{(j)}=P(g_i|(a_l){_{l=1}^j})$, where $a_l$ is the attribute associated with $q_l$ and $r_l$.


Assume that the attribute $a_k$ has $M_k$ distinct values, denoted by ${\bf V_k}=\left\{v_{k,m} |m=1,2,...,M_k\right\}$. We also use $G^{(j)}$ to denote the candidate set of goals for DS-state $S^{(j)}$. Thus, $P_i^{(j)}=0, \forall i\in G,i\notin G^{(j)}$. Therefore, the entropy of state $S^{(j)}$ can be calculated on the current goal set $G^{(j)}$:
\begin{equation}
H^{(j)}	= H^{(j)}(G) = H^{(j)}(G^{(j)})
	= -\sum\limits_{i\in G^{(j)}} P_i^{(j)} {\rm log} P_i^{(j)}.
\label{EMDMEQ1}
\end{equation}
For DS-state $S^{(j)}$, we denote the probabilities of each value of the attribute $a_k$ by $\{P_{k,m}^{(j)} |m=1,2,...,M_k\}$. Then, we use $G_{k,m}^{(j)}$ to denote a subset of the current goal set $G^{(j)}$, whose elements have values of the attribute $a_k$ that are equal to $v_{k,m}$. Then, we have:
\begin{equation}
\sum\limits_{i\in G_{k,m}^{(j)}} P_i^{(j)}=P_{k,m}^{(j)}.
\label{EMDMEQ2}
\end{equation}
In this case, if we can learn that the user's intended value of the attribute $a_k$ is $v_{k,m}$ through one interaction, then the entropy of the remaining goals is:
\begin{equation}
H_{k,m}^{(j)}(G^{(j)}) = -\sum\limits_{i \in G_{k,m}^{(j)}} \frac{P_i^{(j)}}{P_{k,m}^{(j)}} {\rm log} (\frac{P_i^{(j)}}{P_{k,m}^{(j)}}).
\label{EMDMEQ3}
\end{equation}
Therefore, the entropy reduction in this turn of interaction is  $H^{(j)}-H_{k,m}^{(j)}(G^{(j)})$.

We also assume that the users of the dialog system are cooperative and knowledgeable and that, when the system asks a question about the attribute $a_k$, the users will answer with a value $v_{k,m}$ for that attribute $a_k$. The probability of $v_{k,m}$ should follows the current marginal distribution $P_{k,m}^{(j)}$ as calculated in Eq. (\ref{EMDMEQ2}). Thus, the expected entropy reduction upon asking a question related to attribute $a_k$ in DS-state $S^{(j)}$ is as below:
\begin{equation}
\begin{aligned}
&E_k(H^{(j)}-H_{k,m}^{(j)}(G^{(j)})) \\
&= \sum_{m=1}^{M_k} P_{k,m}^{(j)}(H^{(j)}-H_{k,m}^{(j)}(G^{(j)})) \\
&= H^{(j)}-\sum_{m=1}^{M_k} P_{k,m}^{(j)}(-\sum_{i \in G_{k,m}^{(j)}} \frac{P_i^{(j)}}{P_{k,m}^{(j)}} {\rm log} (\frac{P_i^{(j)}}{P_{k,m}^{(j)}})) \\
&= H^{(j)}-\sum_{m=1}^{M_k}(-\sum_{i \in G_{k,m}^{(j)}} P_i^{(j)} {\rm log} (\frac{P_i^{(j)}}{P_{k,m}^{(j)}})) \\
&= H^{(j)}-\sum_{m=1}^{M_k}(-\sum_{i \in G_{k,m}^{(j)}} P_i^{(j)} {\rm log} (P_i^{(j)} + P_i^{(j)} {\rm log}(P_{k,m}^{(j)})) \\
&= H^{(j)}-\sum_{m=1}^{M_k}(-\sum_{i \in G_{k,m}^{(j)}} P_i^{(j)} {\rm log} (P_i^{(j)}))-	\\
& \hspace{2em} \sum_{m=1}^{M_k}(\sum_{i \in G_{k,m}^{(j)}}P_i^{(j)} {\rm log}(P_{m,k}^{(j)})) \\
&= H^{(j)}+\sum_{i=1}^{G^{(j)}} P_i^{(j)} {\rm log}(P_i^{(j)})-\sum_{m=1}^{M_k} {\rm log}(P_{k,m}^{(j)})(\sum_{i \in G_{k,m}^{(j)}}P_i^{(j)}) \\
&= H^{(j)}-H^{(j)}(G^{(j)})-\sum_{m=1}^{M_k} {\rm log}(P_{k,m}^{(j)})P_{k,m}^{(j)} \\
&= H_k^{(j)}.
\end{aligned}
\label{EMDMEQ4}
\end{equation}

$H_k^{(j)}$ represents the entropy of the attribute $a_k$ in the current DS-state $S^{(j)}$. Therefore, the expected value of the entropy reduction is equivalent to that attribute's entropy on the goal set in the DS-state $S_j$. Consequently, the dialog system should generate a question relevant to the attribute with the maximum entropy, with the hope that this question can achieve the maximum entropy reduction. When all questions are generated in this way, the dialog process can be robust and efficient. We refer to this strategy as Entropy Minimization Dialog Management (EMDM).

\begin{figure}[h]
\centering
\includegraphics[width=2.8in,height=2.8in]{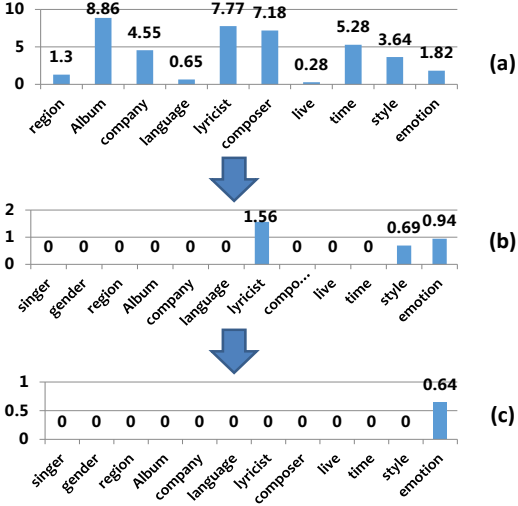}
\caption{An example of entropy evolution using the EMDM strategy.}
\label{FigEntropyChange}
\end{figure}

To illustrate the proposed EMDM strategy, we provide an example in Fig. \ref{FigEntropyChange}. Suppose the goal song is "Under The Moonlight" by the singer "Maggie Chiang" on the album "My room".
The beginning of the dialog may be as follows:
\\ \hspace*{1em} \emph{System: What can I do for you?}
\\ \hspace*{1em} \emph{  User: I'd like to listen to a song.} \\
Initially, based on all songs in the initial goal set and their attributes, the system begins with the marginal distributions of the 12 attributes and their corresponding entropies, shown in Fig. \ref{FigEntropyChange}a. It is evident that the entropy of the attribute \emph{Album} is higher than others. Therefore, the system chooses this attribute to generate a question to obtain a response from the user:
\\ \hspace*{1em} \emph{System: On which album does the song appear?}
\\ \hspace*{1em} \emph{  User: The song is on the album "My Room"} \\
After the album information has been collected, there are 9 songs remaining in the goal set. The entropy of the \emph{Album} attribute drops to 0, and the entropies of the other attributes are also updated, as shown in Fig. \ref{FigEntropyChange}b. The \emph{Lyricist} attribute is now the one with the maximum entropy of 1.56. Therefore, the next question and the corresponding response are as follows:
\\ \hspace*{1em} \emph{System: Who is the lyricist for that song?}
\\ \hspace*{1em} \emph{  User: Peggy Hsu} \\
Based on the lyricist information, there are only 3 candidate songs remaining. The entropies of these attributes are updated as indicated in Fig. \ref{FigEntropyChange}c. Then, the \emph{Emotion} attribute has the maximum entropy of 0.64, and the subsequent interaction between the system and the user is as follows:
\\ \hspace*{1em} \emph{System: What about the emotion?}
\\ \hspace*{1em} \emph{  User: It's a little bit sad.} \\
Finally, we obtain the goal song "Under The Moonlight". Using the EMDM strategy, 3 turns of interaction are required to get the final result. Based on the back-end database and the entropy minimization algorithm, we achieve the goal in a minimal number of dialog turns.

If the user offers a different response, the dialog process will change. For the same goal song, we provide another example as follows:
\\ \hspace*{1em} \emph{System: What can I do for you?}
\\ \hspace*{1em} \emph{  User: I would like a song by Maggie Chiang}
\\ \hspace*{1em} \emph{System: What kind of song?}
\\ \hspace*{1em} \emph{  User: A popular one}
\\ \hspace*{1em} \emph{System: Who is the lyricist of that song?}
\\ \hspace*{1em} \emph{  User: Peggy Hsu}
\\ \hspace*{1em} \emph{System: On what album does that song appear?}
\\ \hspace*{1em} \emph{  User: It is on My Room.} \\
Only one candidate song remains after 4 turns of interaction. It demonstrates that the dialog manager can adapt to the situation and still achieve a minimal number of dialog turns.

\section{DS-state Evolution With Multiple SLU Candidates}
\label{SecDSS}
As mentioned in Section \ref{SecProbFrame}, we use an input processing model to address the user's input utterances. The primary task of this input model is, beginning from the current DS-state $S^{(j)}$, to update the distribution over the goal set, $P_i^{(j+1)}=P(g_{i+1}|S^{(j+1)},D$) and reach the next DS-state, $S^{(j+1)}$. The examples presented in Section \ref{SecEMDM} illustrate that the proposed EMDM strategy is very effective in the ideal situation. However, errors will be unavoidable in the ASR and NLU modules. We believe that a combination of DS-states and EMDM is capable of coping with these errors and help to guide the dialog process in a robust and efficient manner.

In the $j^{th}$ interaction, in the DS-state $S^{(j)}$ with the probability, $\{P_i^j|i \in G^{(j)}\}$, and the next DS-state will be $S^{(j+1)}$ with the probability, $\{P_i^{(j+1)}|i \in G^{(j+1)}\}$. For question $q^{(j)}$ issued by the dialog system, each candidate in the SLU results from the user's answer has a posterior probability, which is estimated as the ratio of the probability of this candidate to the whole search space \cite{kemp1997estimating, abdou2004beam, jiang2005confidence}. These probabilities can be treated as a confidence measure of the corresponding song in the goal set, denoted by $\{C_i^{(j)}|i \in G^{(j)}\}$. Then, we update the distribution over the goal set using the rule proposed in \cite{wang2013simple}. Thus, in the $j^{th}$ turn, the probabilities of the songs in the goal set are evaluated as follows:


\begin{equation}
\begin{aligned}
& P_i^{(j+1)}=1-(1-P_i^{(j)})(1-C_i^{(j)}), \hspace{0.3em} \forall i, C_i^{(j)}>0 \\
& P_i^{(j+1)}=(1-\sum\limits_{k\in G^{(j)}} C_k^{(j)})P_i^{(j)}, \hspace{2em} \forall i, C_i^{(j)}=0 .
\label{DSSEQ1}
\end{aligned}
\end{equation}


In this way, we can extract more information from the multiple candidates obtained by the SLU module based on the user's inputs. This flexibility facilitates the dialog system to efficiently cope with unavoidable ASR and NLU errors.

\section{Experiments and Result Analysis}
\label{SecExperiment}

\subsection{Dialog Management Experiments in Ideal Settings}
Using real dialogs between different users and a multiple-interaction dialog system is a good way to test the proposed DM strategies. However such a test is also very expensive to conduct. Therefore, for this series of experiments, we first designed a simulation scenario with a knowledgeable and cooperative user. For each dialog, the simulated user first chose a song from the back-end database as the goal, and the dialog system was given no information concerning the chosen song. When the dialog process began, in each interaction, the system asked one question about one particular attribute (e.g., ''Who is the singer of the song''), and the simulated user answered in a cooperative way. Then, the system updated the goal set based on its understanding of the user's utterance. This process repeated until one of the following three conditions was met:
\\ 1) only one song remained in the candidate set,
\\ 2) the entropies of all 12 attributes dropped to 0, or
\\ 3) all attributes had been inquired about by the system. \\
The dialog process concluded when the system returned a song to the user. Because there was no error in the results of parsing the user's utterances, the dialog could always achieve the correct target. Therefore, we used only the average number of turns required to reach an answer as an evaluation metric.



We compared the proposed EMDM strategy with three others, sequential, random, and DSDM discussed in \cite{polifroni2006learning}. The sequential strategy chose questions in a fixed order, one by one, consistent with Table \ref{12ATTRs} until a termination condition was met. The random strategy chose attributes in a random order. The DSDM method was an approximate entropy-based method. It assumed that attributes with more distinct values should have higher entropies, and that such attributes should be addressed first. We conduct the experiments in two different settings. First, we do not have any knowledge about which song is preferred by the users, and therefore every song in the database share the same prior probability. In this uniform setting case, every song was evaluated once with the four DM strategies. Second, we do have a prior distribution over the database, which is obtained from the users' dialog history. In this sampling setting case, 500,000 songs were sampled from the database according to the prior distribution and evaluated with different strategies. The average number of dialog turns required for each strategy is listed in Table \ref{DMResults}.

\begin{table}[h]
\renewcommand{\arraystretch}{1.3}
\caption{Average numbers of rounds of dialog for different strategies}
\label{DMResults}
\centering
\begin{tabular}{|c|c|c|}
\hline
\textbf{Strategy} & \textbf{uniform} & \textbf{sampling}\\
\hline
\hline
\emph{Sequential} & 9.298 & 8.306 \\
\hline
\emph{Random} & 8.297 & 7.159 \\
\hline
\emph{DSDM} & 3.330 & 3.223 \\
\hline
\emph{\textbf{EMDM}} & 3.309 & 3.065 \\
\hline
\end{tabular}
\end{table}

\begin{figure}[h]
\centering
\subfloat[]{\includegraphics[width=1.7in,height=0.7in]{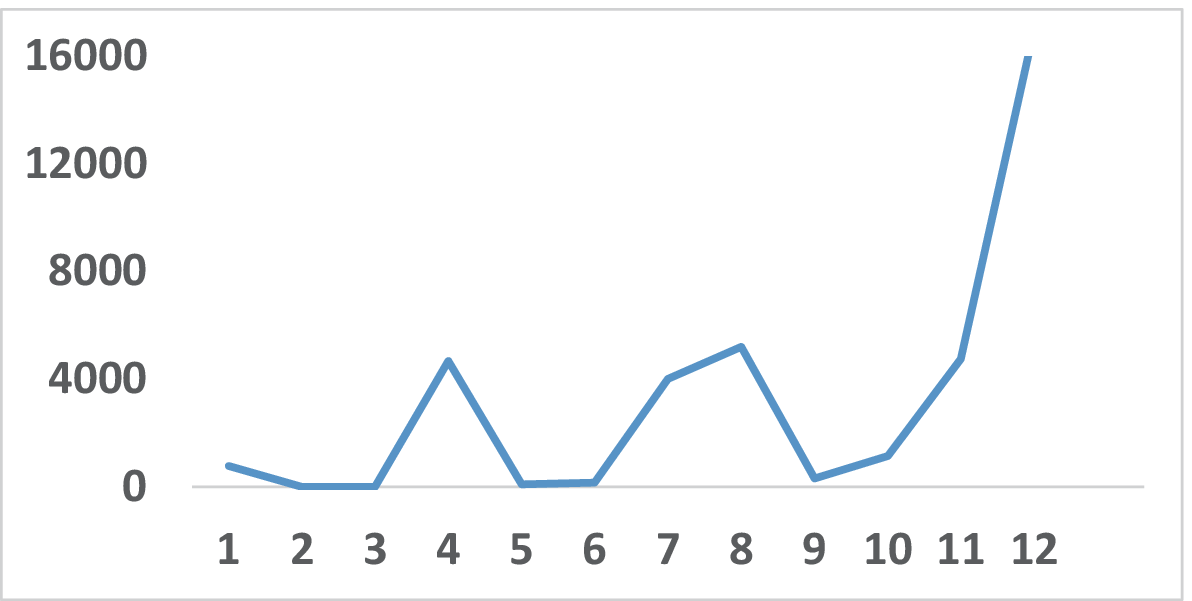}%
\label{SEQ}}
\subfloat[]{\includegraphics[width=1.7in,height=0.7in]{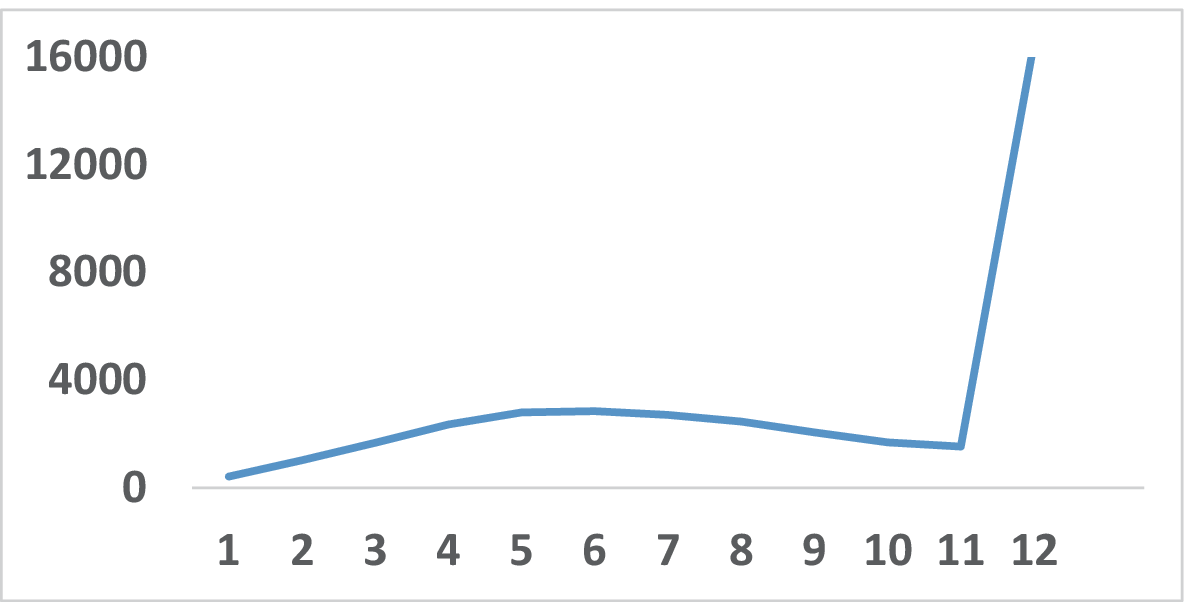}%
\label{RAN}}
 \hfil
\subfloat[]{\includegraphics[width=1.7in,height=0.7in]{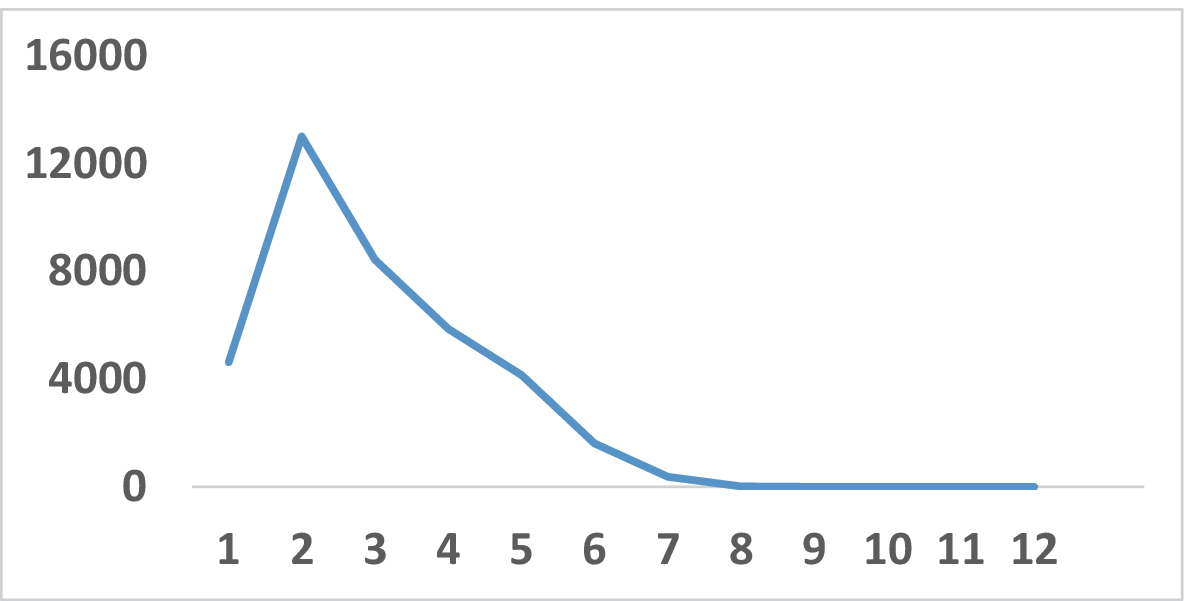}%
\label{ProEnt}}
\subfloat[]{\includegraphics[width=1.7in,height=0.7in]{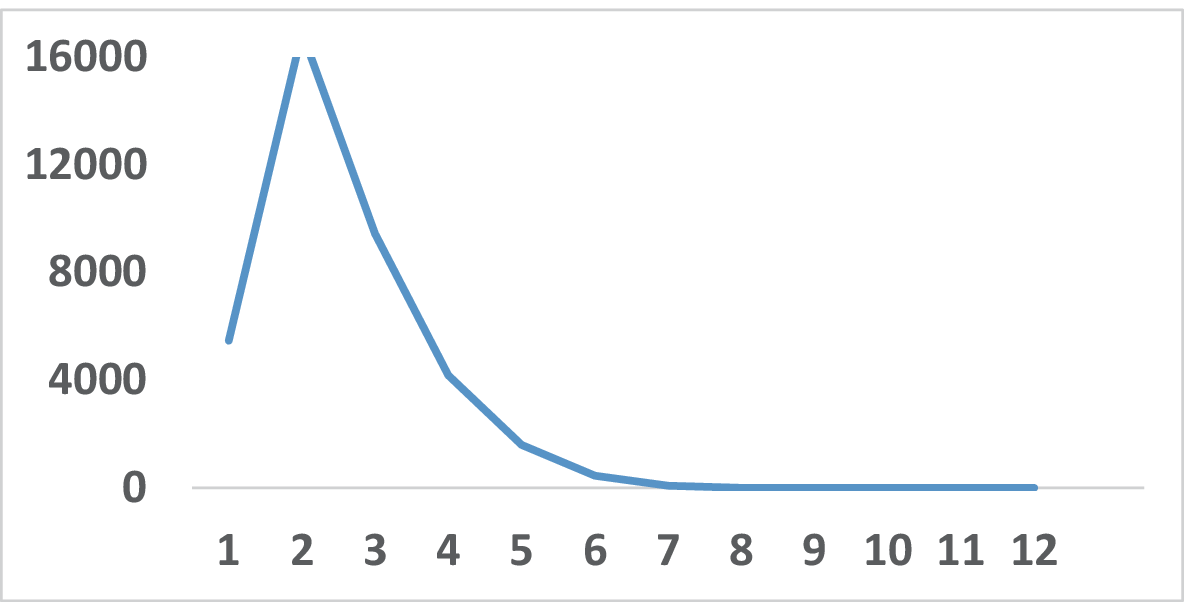}%
\label{ENT}}
\caption{Histograms of the number of dialog turns required by the four DM strategies: (a) sequential; (b) random; (c) DSDM; and (d) EMDM.}
\label{DMResultsFig}
\end{figure}

With the uniform setting in the middle column, the sequential and random strategy needed 9.3 and 8.3 dialog turns to achieve the target song on average. With the sampling setting in the right column, 8.3 and 7.2 dialog turns were needed by these two strategies, respectively. Comparing sequential and random strategies, shown in Figs. \ref{SEQ} and \ref{RAN}, we found that the performance curve of the random strategy is much smoother than that of the sequential strategy. For the latter, the attributes were addressed in a fixed order. If we choose another order different from \ref{12ATTRs}, the performance must be changed. Therefore, the performance curve of sequential strategy reflects the characteristics of the database, whereas the performance of random strategy could be more stable. For these two strategies, unless the system reached a state in which only one candidate song remained after several turns, the dialog process did not end before queries related to all 12 attributes had been issued. Therefore, a large percentage of the dialogs required 12 turns of interaction. In fact, in some cases, if information on a certain attribute had already been acquired, it would be a waste to ask about certain other attributes. For example, if we already knew the singer of the goal song, then inquiring about the gender or region of the singer would be of no benefit. In contrast, the DSDM and EMDM strategies were both based on entropy analysis over the goal set. For a certain dialog, when all attributes of the remaining candidates had the same values, the entropy of the goal set became zero. Even if there were more than one song in the goal set, the dialog would stop. In such a case, fewer than 12 interaction turns were inquired. Furthermore, based on the analysis, these two entropy-based methods could choose informative questions to ask the users, giving a large reduction in the number of turns. As a result, the peaks in the performance curves on the upper right sides in Figs. \ref{DMResultsFig}(a) and \ref{DMResultsFig}(b) moved to the lower left sides of the curves in Figs. \ref{DMResultsFig}(c) and \ref{DMResultsFig}(d) showing fewer dialog turns.




The DSDM strategy chose the attribute with the maximum number of distinct values which might not be the most informative one. For example, if a set of songs corresponded to many distinct composers, but most of the songs were composed by only one of them, then the \emph{composer} attribute might not be the most effective choice. The EMDM strategy was able to chose the most informative attribute about which to ask and therefore performed even better than the DSDM method. However, the difference was not that much. To explain that results, a more detailed comparison is listed in Table \ref{DMDetailResults}.

\begin{table}[h]
\renewcommand{\arraystretch}{1.3}
\caption{Comparison of the EMDM and DSDM Strategies}
\label{DMDetailResults}
\centering
\begin{threeparttable}[b]
\begin{tabular}{|c|c|c|c|c|}
\hline
\textbf{setting} & $\#E\tnote{1} < \#D\tnote{2}$ & $\#E\tnote{1} = \#D\tnote{2}$ & $\#E\tnote{1} > \#D\tnote{2}$ & \textbf{total}\\
\hline
\hline
\ Uniform & 4.09\% & 93.68\% & 2.23\% & 38117 \\
\hline
\ Sampling & 15.38\% & 82.75\% & 1.87\% & 500,000 \\
\hline
\end{tabular}
\begin{tablenotes}
        \footnotesize
        \item[1] $\#E$: number of dialog turns required by EMDM strategy
        \item[2] $\#D$: number of dialog turns required by DSDM strategy
\end{tablenotes}
\footnotesize
\end{threeparttable}
\end{table}



Because the DSDM and EMDM strategies are both based on entropy analysis, so in most situations, they make the same choice in the question to be asked next. However even with the "Uniform" setting shown in the top row, we still see that the chance for EMDM (4.09\%) to perform better about two times than DSDM (2.23\%). With the "Sampling" setting in the bottom row, the chance for EMDM (15.38\%) to be better is more than 8 times than DSDM (1.87\%), i.e., the proposed EMDM strategy can manage the dialog process in a more efficient and flexible manner than other competing strategies.

\subsection{Dialog Management Experiments with SLU Errors}
In the proposed probabilistic representation of the dialog process, the dialog DS-states are characterized by the distributions over the goal set, and can be updated based on the outputs of the SLU module using not only the top candidate, but also a set of multiple candidates in order to effectively cope with the potential ASR and NLU errors. Using the same back-end database we conducted a series of online tests with six real users. Focusing on the subset of 60 specific songs, we asked each subject to find 10 songs of the subset using the four tested strategies, sequential, random, DSDM, and EMDM. This yielded 60 test cases for each strategy.

Based on the multiple candidates provided by the SLU module, Eq. (\ref{DSSEQ1}) was used to update the current DS-state. Because the output candidates from the SLU module were each associated with a certain confidence measure, we could never obtain a goal song with a total certainty. Therefore, we changed the dialog termination conditions as follows:
\\ 1) one candidate song was dominant in probability, for example, more than 80\% percent;
\\ 2) all attributes had been inquired about by the system, or;
\\ 3) no candidates remained in the goal set.

\begin{table}[h]
\renewcommand{\arraystretch}{1.3}
\caption{Performance of DM Strategies with SLU Errors}
\label{MultiCandPfm}
\centering

\begin{threeparttable}[b]
\begin{tabular}{|c||c|c|c|c|}
\hline

\textbf{Dialog} & \textbf{ASR Accu} & \textbf{SLU Accu} & \textbf{Dialog} & \textbf{Turns of} \\
\textbf{Strategy} & \textbf{(in \%)} & \textbf{(in \%)} & \textbf{Success Rate} & \textbf{Interaction} \\
\hline
\hline
\emph{Sequential} & 90.9 & 90.6 & 50.0\%  & 8.75 \\
\hline
\emph{Random} & 89.3 & 88.5 &  61.7\% & 6.23  \\
\hline
\emph{DSDM-Top5} & 84.5 (88.7)\tnote{1} & 82.7 (88.4)\tnote{1} & 80.0\% & 5.73  \\
\hline
\emph{\textbf{EMDM-Top5}} & 85.4 (89.2)\tnote{1} & 83.5 (88.8)\tnote{1} & \textbf{86.7\%} & \textbf{5.17}  \\
\hline
\end{tabular}
\begin{tablenotes}
        \footnotesize
        \item[1] \emph{Prediction accuracy for top 5 output candidates shown in parentheses}
\end{tablenotes}
\end{threeparttable}
\end{table}

As shown in Table \ref{MultiCandPfm}, the unavoidable ASR and NLU errors had a manifest impact on the performance of the four DM strategies. In the top row when the system asked the questions in a fixed order, only 50\% of the dialogs finished successfully within 8.75 turns of the interaction on average. When random questions were asked by the system, the success rate was 61.7\% and an average of 6.23 dialog turns was required. To handle these ASR and NLU errors more effectively, we utilized the top 5 SLU candidates to update the DS-states of the dialogs. In this case, the DSDM strategy required 5.73 dialog turns to achieve an 80\% success rate. For the proposed EMDM strategy, the success rate was 86.7\% and only 5.17 dialog turns were required on average, corresponding to the best performance among all tested strategies.

\begin{figure}[h]
\centering
\includegraphics[width=2.5in,height=1.2in]{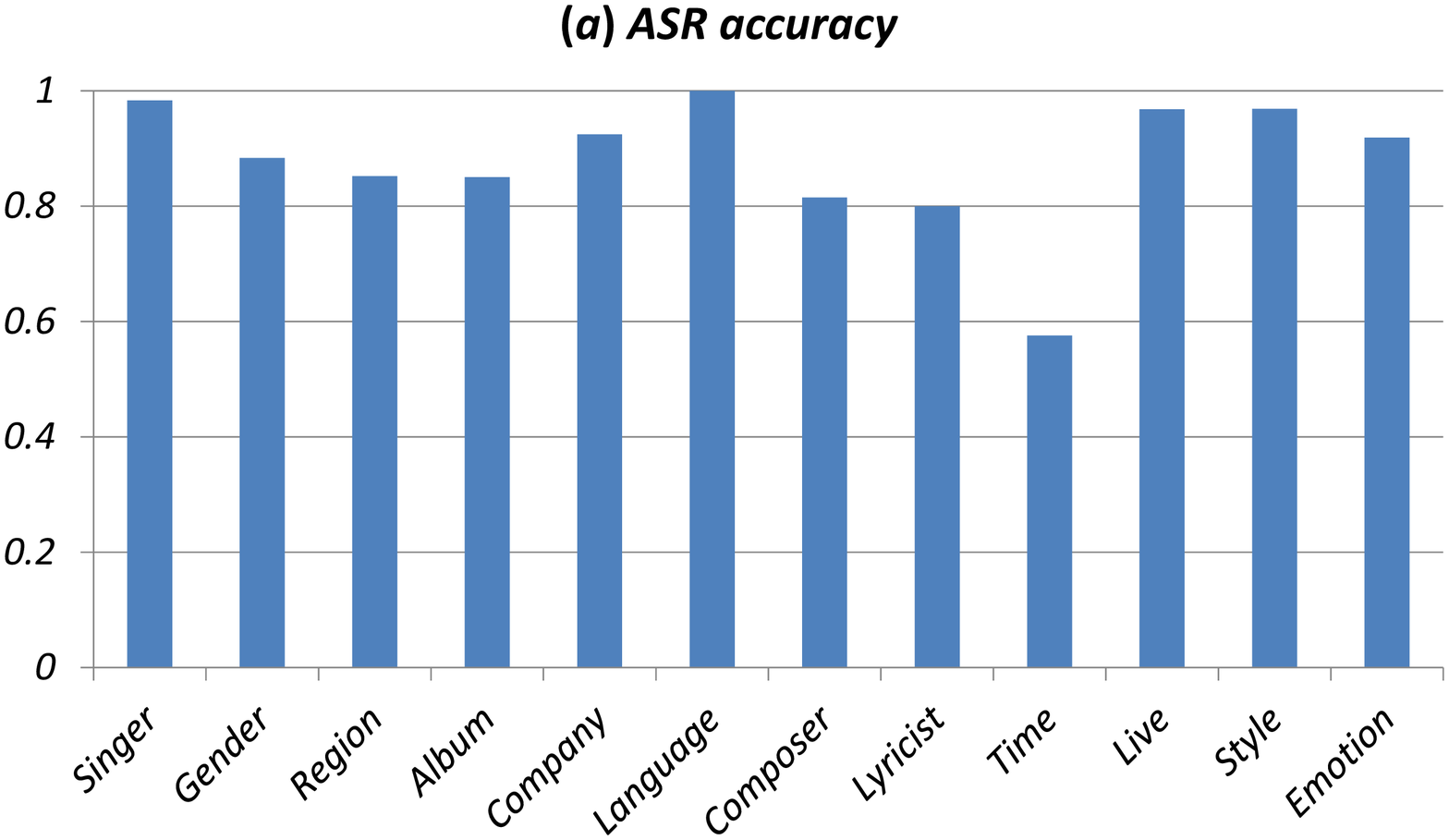}%
\label{ASRAcc}
\hfil
\includegraphics[width=2.5in,height=1.2in]{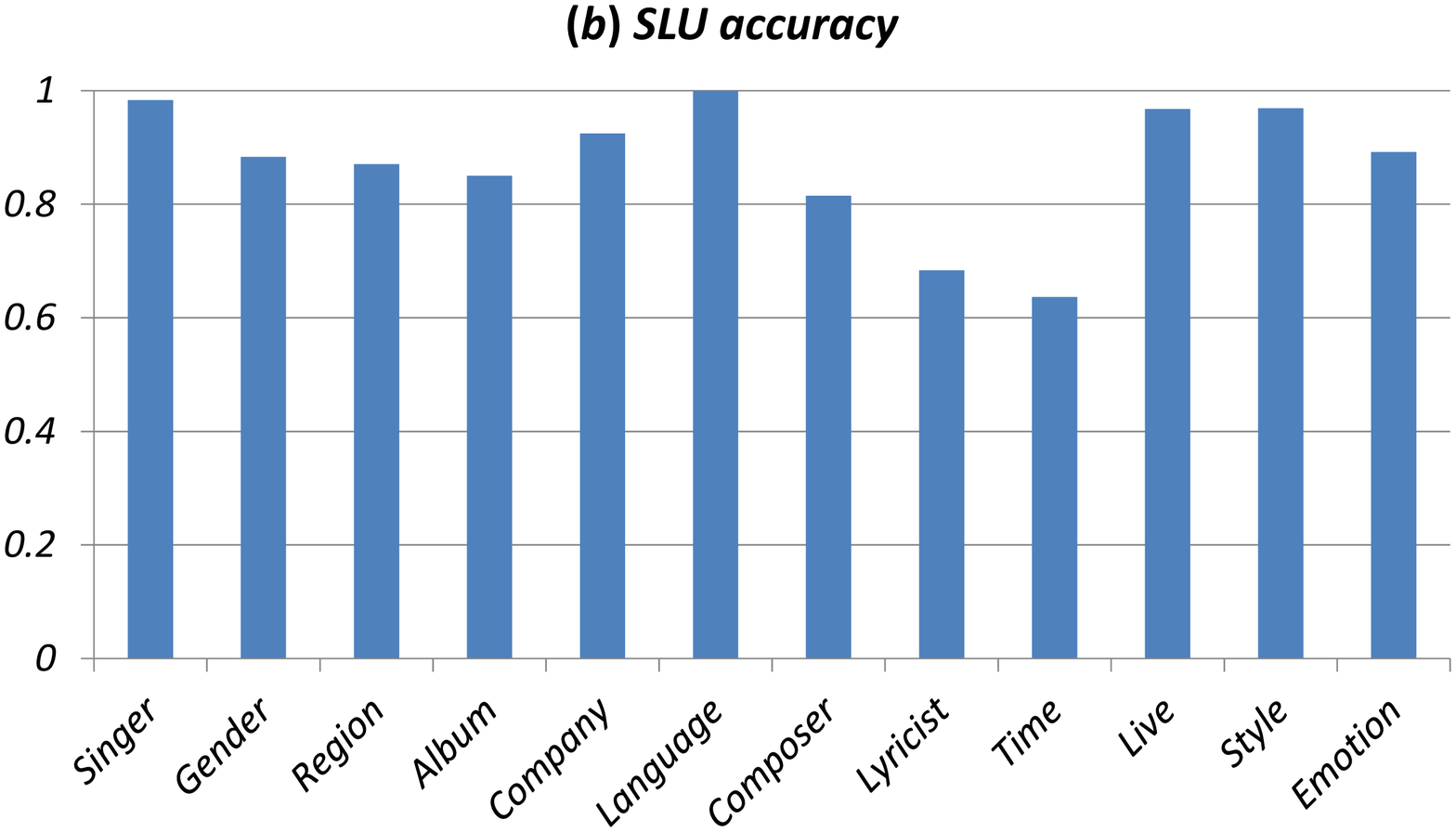}%
\label{SLUAcc}
\hfil
\caption{The ASR and SLU accuracies of user's utterance that corresponding to questions about different attributes (a)\emph{ASR accuracy distribution over the attributes}; (b) \emph{SLU accuracy distribution over the attributes}.}
\label{UtteranceFig}
\end{figure}

\begin{figure}[h]
\centering
\includegraphics[width=2.5in,height=1.1in]{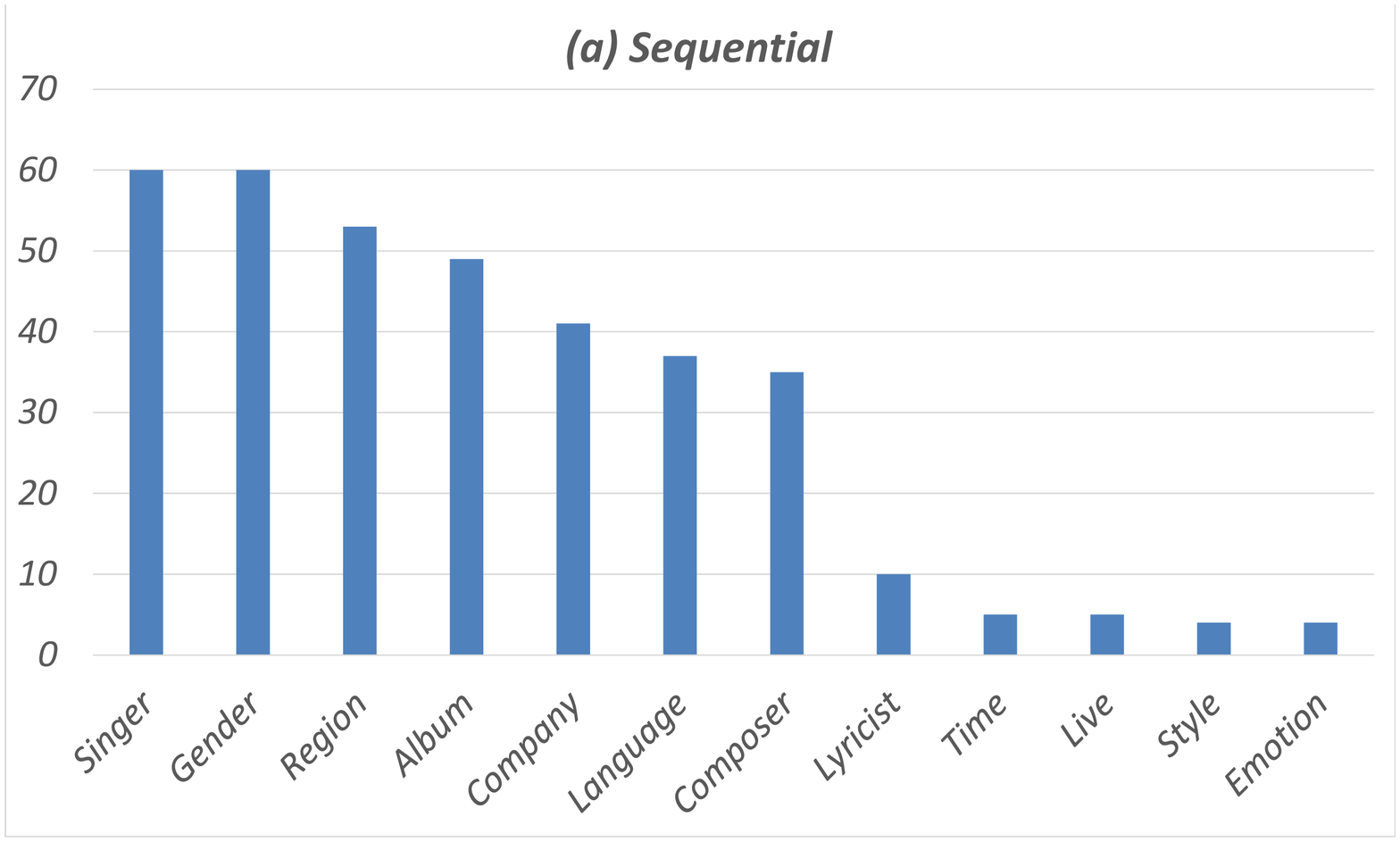}%
\label{SeqStategy}
\hfil
\includegraphics[width=2.5in,height=1.1in]{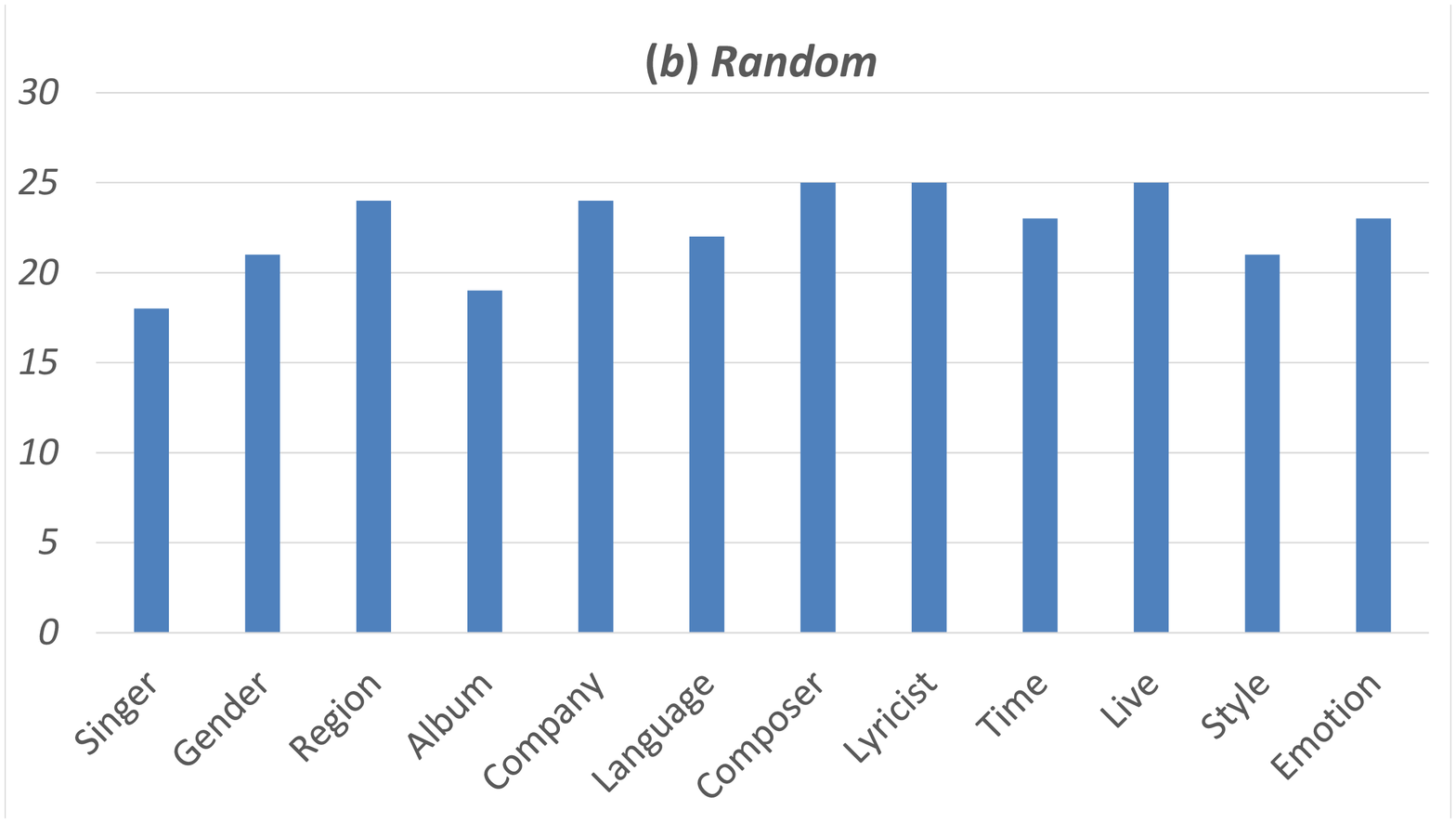}%
\label{RandomStategy}
\hfil
\includegraphics[width=2.5in,height=1.1in]{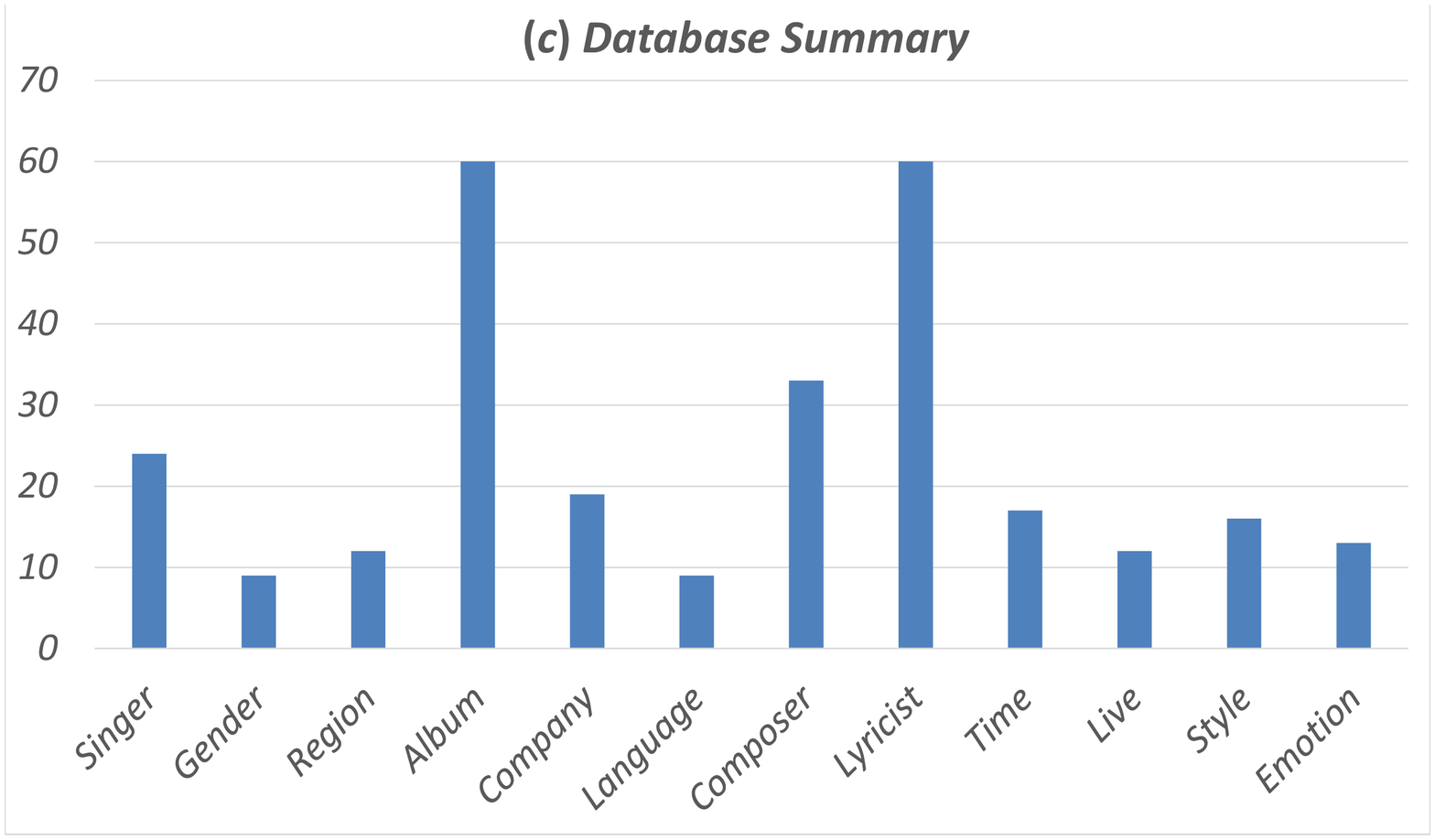}%
\label{DSDMStrategy}
\hfil
\includegraphics[width=2.5in,height=1.1in]{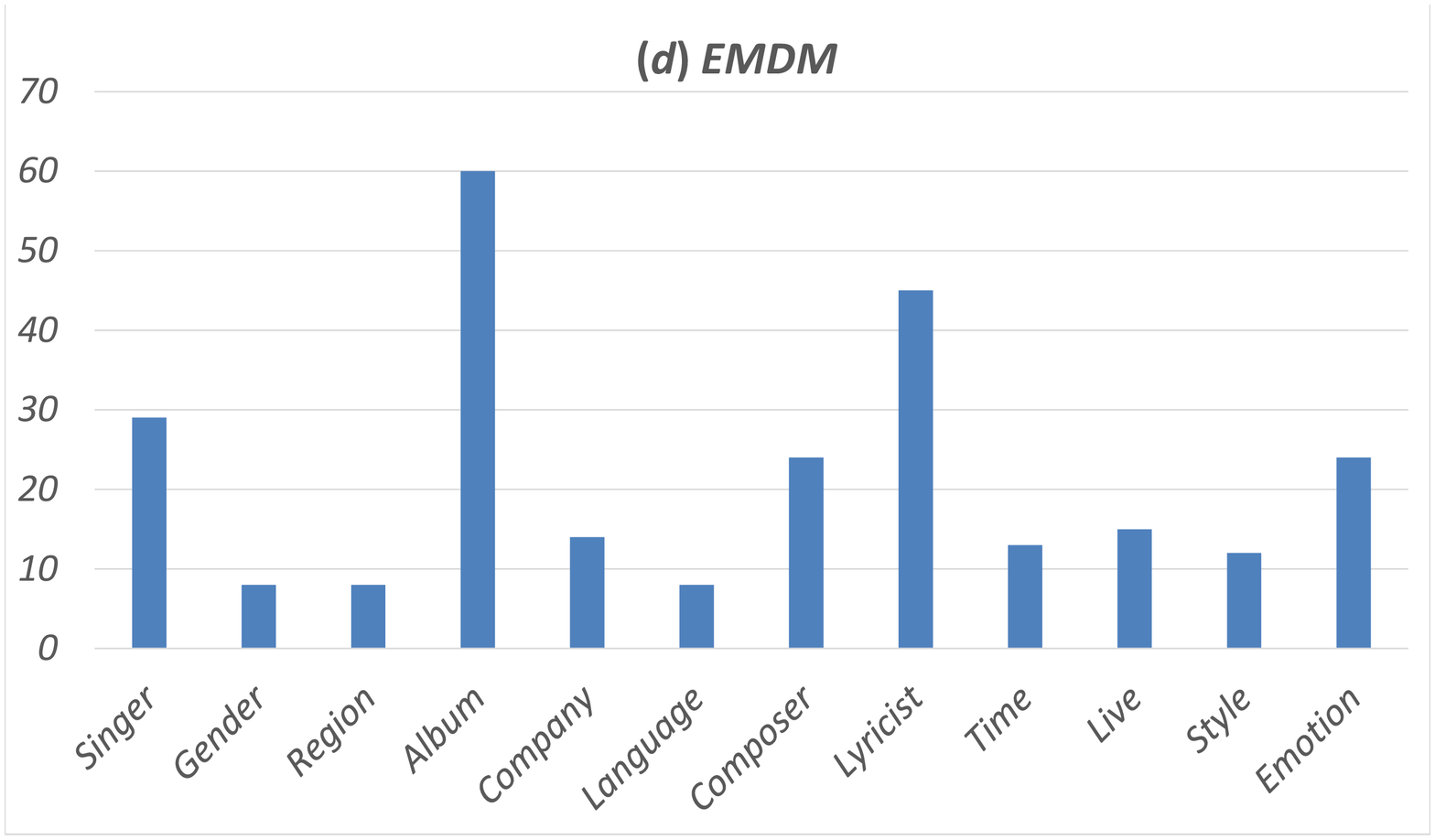}%
\label{ EMDMStategy}
\hfil
\caption{Distributions of the presented questions over the attributes in the four DM strategies: (a)\emph{sequential}; (b) \emph{random}; (c)\emph{DSDM}; and (d) \emph{EMDM}. }
\label{StrategyFig}
\end{figure}

The reason that different DM strategies exhibited different ASR and SLU accuracies could come from the fact that different questions were asked in different situations with different expected vocabulary complexities. Indeed, this is one of the key difficulties in comparing multiple-interaction dialog systems. Even for the same target and the same starting point, the dialog follows different paths if the system choose a different action on any given dialog turn. Therefore, for the different DM strategies, the ASR and SLU performances were evaluated on different test data sets. For questions about different attributes, the users' answers yielded different ASR and SLU errors. As shown in Fig. \ref{UtteranceFig}, certain attributes, such as \emph{Language}, \emph{Singer}, \emph{Style} and \emph{Emotion}, tended to yield better accuracies, whereas for other attributes, such as \emph{Time}, \emph{Lyricist}, \emph{Composer} and \emph{Region}, the performances were much worse. The DM strategies also followed different distributions regarding which questions they chose to ask as shown in Fig. \ref{StrategyFig}. Because the sequential strategy followed a fixed order, later attributes were less frequently inquired about because more dialogs terminated before reaching them as shown in Fig. \ref{StrategyFig}(a). On the other hand the random strategy followed a uniform distribution in inquiring about the attributes as shown in Fig. \ref{StrategyFig}(b). From Fig. \ref{UtteranceFig} and Fig. \ref{StrategyFig}, it is evident that the sequential DM strategy demonstrated slightly better ASR and SLU performances because certain attributes with high recognition errors, such as \emph{Lyricist}, were rarely asked about. As shown in Fig. \ref{StrategyFig}(c) and Fig. \ref{StrategyFig}(d), the DSDM and EMDM strategies presented a large number of questions related to the attributes \emph{Album} and \emph{Lyricist}, which led to worse ASR and SLU accuracies as shown in the two bottom rows in Table \ref{MultiCandPfm}.

For the sequential and random strategies, when the dialog system received a user's input utterances, the system used the SLU results as an additional constraint. The dialog process stopped when a termination conditions was met. Sometimes, the single remaining candidate song was different from the correct target, and sometimes there was no song left at all, which significantly degraded the dialog turn performances as shown in the top two rows in Table \ref{MultiCandPfm}. Although the top-1 performances of the DSDM and EMDM strategies were not high, the SLU model also gave top 5 results (shown in parentheses together with the top-1 ASR and SLU accuracies), which yielded much better prediction performances. Furthermore, by virtue of the DS-state updating method, we could make a good use of these multiple candidates and always increase the probability of a presence of the correct targets in the candidate set. Meanwhile, the entropy-based approaches could also identify more informative questions to ask as the next system action. The EMDM strategy obtained the best performance in Table \ref{MultiCandPfm}, as expected, because it took into account the distributions over the goal set, whereas the DSDM approach simply counted how many distinct values remained in those relevant attributes.

\section{Conclusion and Future Work}
\label{SecConclusion}
In this paper, a probabilistic framework for representing spoken dialog systems is presented to model dialog interactions in which the user and the system work collaboratively to achieve users' goals  in a robust and efficient manner. To describe the dialog process more accurately and efficiently, we define a new dynamic stochastic state which is characterized by a distribution over the goal set. Furthermore, an entropy minimization dialog management strategy is proposed to minimize the entropy of the goal set as rapidly as possible.

Based on the framework, we propose an efficient and effective solution consisting of two components, the updating of the DS-states based on multiple candidates from the SLU module and the execution of the EMDM strategy. Our experiments demonstrate that we could achieve the highest dialog success rate in the fewest turns of interaction using this solution. A Song-On-Demand task related to the retrieval of songs from a real-world music database was used in to evaluate the proposed solution. In a simulation situation with no ASR and SLU errors, we show that the proposed collaborative EMDM strategy yields the best results in terms of the least number of interaction turns required to accomplish a task. The comparison of four different DM strategies indicate that, entropy-based DM strategies are more efficient and effectively than non-entropy based DM. Moreover, with goal set distributions in EMDM, the efficiency is also better than those without them in DSDM. Furthermore, in practical scenario testing, with potential ASR and SLU errors, the proposed DS-states and EMDM strategy combination effectively exploits multiple candidates and therefore achieves the best dialog success rate and the minimum number of average dialog turns.

So far in this study, it was assumed that the user has a full knowledge of the attributes associated with each goal. However, in a real-world dialog scenario, only partial knowledge is available. To address this issue, the model component $p(r^{(j+1)}|q^{(j+1)},D)$ mentioned in Section \ref{SecProbFrame} can be integrated into the probabilistic framework. We believe that the proposed dynamic stochastic states and entropy minimization dialog management strategy provide an efficient approach to designing and implementing dialog systems. Moreover, the framework is amenable to the integration of additional factors that may be relevant in real-world dialog problems, such as a user's preferences with regard to different goals, the characteristics of a user's behavior in different scenarios, and even changes of a user's mind.


%



\section*{Acknowledgment}
 The first two authors were funded for this work by the National Natural Science Funds of China under Grant 61170197, the Electronic Information Industry Development Fund of China under project ”The R\&D and Industrialization on Information Retrieval System Based on Man-Machine Interaction with Natural Speech”, the National High-Tech. R\&D Program of China (863 Program) under Grant 2012AA011004, and the Planned Science and Technology Project of Tsinghua University under Grant 20111081023.


\ifCLASSOPTIONcaptionsoff
  \newpage
\fi


\begin{thebibliography}{10}
\providecommand{\url}[1]{#1}
\csname url@samestyle\endcsname
\providecommand{\newblock}{\relax}
\providecommand{\bibinfo}[2]{#2}
\providecommand{\BIBentrySTDinterwordspacing}{\spaceskip=0pt\relax}
\providecommand{\BIBentryALTinterwordstretchfactor}{4}
\providecommand{\BIBentryALTinterwordspacing}{\spaceskip=\fontdimen2\font plus
\BIBentryALTinterwordstretchfactor\fontdimen3\font minus
  \fontdimen4\font\relax}
\providecommand{\BIBforeignlanguage}[2]{{%
\expandafter\ifx\csname l@#1\endcsname\relax
\typeout{** WARNING: IEEEtran.bst: No hyphenation pattern has been}%
\typeout{** loaded for the language `#1'. Using the pattern for}%
\typeout{** the default language instead.}%
\else
\language=\csname l@#1\endcsname
\fi
#2}}
\providecommand{\BIBdecl}{\relax}
\BIBdecl

\bibitem{jokinen2009spoken}
K.~Jokinen and M.~McTear, ``Spoken dialogue systems,'' \emph{Synthesis Lectures
  on Human Language Technologies}, vol.~2, no.~1, pp. 1--151, 2009.

\bibitem{lee2000natural}
C.-H. Lee, B.~Carpenter, W.~Chou, J.~Chu-Carroll, W.~Reichl, A.~Saad, and
  Q.~Zhou, ``On natural language call routing,'' \emph{Speech Communication},
  vol.~31, no.~4, pp. 309--320, 2000.

\bibitem{seneff2000dialogue}
S.~Seneff and J.~Polifroni, ``Dialogue management in the mercury flight
  reservation system,'' in \emph{Proceedings of the 2000 ANLP/NAACL Workshop on
  Conversational systems-Volume 3}.\hskip 1em plus 0.5em minus 0.4em\relax
  Association for Computational Linguistics, 2000, pp. 11--16.

\bibitem{goddeau1994galaxy}
D.~Goddeau, E.~Brill, J.~R. Glass, C.~Pao, M.~Phillips, J.~Polifroni,
  S.~Seneff, and V.~W. Zue, ``Galaxy: A human-language interface to on-line
  travel information,'' in \emph{Third International Conference on Spoken
  Language Processing}, 1994.

\bibitem{weng2007chat}
F.~Weng, B.~Yan, Z.~Feng, F.~Ratiu, M.~Raya, B.~Lathrop, A.~Lien, S.~Varges,
  R.~Mishra, F.~Lin \emph{et~al.}, ``Chat to your destination,'' in \emph{Proc.
  of the 8th SIGDial workshop on Discourse and Dialogue}, 2007, pp. 79--86.

\bibitem{rabiner1993fundamentals}
L.~R. Rabiner and B.-H. Juang, \emph{Fundamentals of speech recognition}.\hskip
  1em plus 0.5em minus 0.4em\relax PTR Prentice Hall Englewood Cliffs, 1993,
  vol.~14.

\bibitem{povey2011kaldi}
D.~Povey, A.~Ghoshal, G.~Boulianne, L.~Burget, O.~Glembek, N.~Goel,
  M.~Hannemann, P.~Motl{\'\i}{\v{c}}ek, Y.~Qian, P.~Schwarz \emph{et~al.},
  ``The kaldi speech recognition toolkit,'' 2011.

\bibitem{hinton2012deep}
G.~Hinton, L.~Deng, D.~Yu, G.~E. Dahl, A.-r. Mohamed, N.~Jaitly, A.~Senior,
  V.~Vanhoucke, P.~Nguyen, T.~N. Sainath \emph{et~al.}, ``Deep neural networks
  for acoustic modeling in speech recognition: The shared views of four
  research groups,'' \emph{Signal Processing Magazine, IEEE}, vol.~29, no.~6,
  pp. 82--97, 2012.

\bibitem{zue2000conversational}
V.~W. Zue and J.~R. Glass, ``Conversational interfaces: Advances and
  challenges,'' \emph{Proceedings of the IEEE}, vol.~88, no.~8, pp. 1166--1180,
  2000.

\bibitem{rosenfield2000two}
R.~Rosenfield, ``Two decades of statistical language modeling: Where do we go
  from here?'' 2000.

\bibitem{huang2001spoken}
X.~Huang, A.~Acero, H.-W. Hon, and R.~Foreword By-Reddy, \emph{Spoken language
  processing: A guide to theory, algorithm, and system development}.\hskip 1em
  plus 0.5em minus 0.4em\relax Prentice Hall PTR, 2001.

\bibitem{mesnil2013investigation}
G.~Mesnil, X.~He, L.~Deng, and Y.~Bengio, ``Investigation of
  recurrent-neural-network architectures and learning methods for spoken
  language understanding.'' in \emph{INTERSPEECH}, 2013, pp. 3771--3775.

\bibitem{bellegarda2014spoken}
J.~R. Bellegarda, ``Spoken language understanding for natural interaction: The
  siri experience,'' in \emph{Natural Interaction with Robots, Knowbots and
  Smartphones}.\hskip 1em plus 0.5em minus 0.4em\relax Springer, 2014, pp.
  3--14.

\bibitem{scheffler1999simulation}
K.~Scheffler and S.~Young, ``Simulation of human-machine dialogues,'' in
  \emph{Proc. ICASSP}.\hskip 1em plus 0.5em minus 0.4em\relax Citeseer, 1999.

\bibitem{young2010hidden}
S.~Young, M.~Ga{\v{s}}i{\'c}, S.~Keizer, F.~Mairesse, J.~Schatzmann,
  B.~Thomson, and K.~Yu, ``The hidden information state model: A practical
  framework for pomdp-based spoken dialogue management,'' \emph{Computer Speech
  \& Language}, vol.~24, no.~2, pp. 150--174, 2010.

\bibitem{williams2007partially}
J.~D. Williams and S.~Young, ``Partially observable markov decision processes
  for spoken dialog systems,'' \emph{Computer Speech \& Language}, vol.~21,
  no.~2, pp. 393--422, 2007.

\bibitem{pieraccini2005we}
R.~Pieraccini and J.~Huerta, ``Where do we go from here? research and
  commercial spoken dialog systems,'' in \emph{6th SIGdial Workshop on
  Discourse and Dialogue}, 2005.

\bibitem{pellom2001university}
B.~Pellom, W.~Ward, J.~Hansen, R.~Cole, K.~Hacioglu, J.~Zhang, X.~Yu, and
  S.~Pradhan, ``University of colorado dialog systems for travel and
  navigation,'' in \emph{Proceedings of the first international conference on
  Human language technology research}.\hskip 1em plus 0.5em minus 0.4em\relax
  Association for Computational Linguistics, 2001, pp. 1--6.

\bibitem{hanna2007promoting}
P.~Hanna, I.~O'neill, C.~Wootton, and M.~Mctear, ``Promoting extension and
  reuse in a spoken dialog manager: An evaluation of the queen's
  communicator,'' \emph{ACM Transactions on Speech and Language Processing
  (TSLP)}, vol.~4, no.~3, p.~7, 2007.

\bibitem{Schatzmann2006Survey}
J.~Schatzmann, K.~Weilhammer, M.~Stuttle, and S.~Young, ``A survey of
  statistical user simulation techniques for reinforcement-learning of dialogue
  management strategies,'' \emph{The Knowledge Engineering Review}, vol.~21,
  no.~02, pp. 97--126, 2006.

\bibitem{levin1998using}
E.~Levin, R.~Pieraccini, and W.~Eckert, ``Using markov decision process for
  learning dialogue strategies,'' in \emph{Acoustics, Speech and Signal
  Processing, 1998. Proceedings of the 1998 IEEE International Conference on},
  vol.~1.\hskip 1em plus 0.5em minus 0.4em\relax IEEE, 1998, pp. 201--204.

\bibitem{levin2000stochastic}
------, ``A stochastic model of human-machine interaction for learning dialog
  strategies,'' \emph{Speech and Audio Processing, IEEE Transactions on},
  vol.~8, no.~1, pp. 11--23, 2000.

\bibitem{young2000probabilistic}
S.~J. Young, ``Probabilistic methods in spoken--dialogue systems,''
  \emph{Philosophical Transactions of the Royal Society of London. Series A:
  Mathematical, Physical and Engineering Sciences}, vol. 358, no. 1769, pp.
  1389--1402, 2000.

\bibitem{barto1998reinforcement}
A.~G. Barto, \emph{Reinforcement learning: An introduction}.\hskip 1em plus
  0.5em minus 0.4em\relax MIT press, 1998.

\bibitem{roy2000spoken}
N.~Roy, J.~Pineau, and S.~Thrun, ``Spoken dialogue management using
  probabilistic reasoning,'' in \emph{Proceedings of the 38th Annual Meeting on
  Association for Computational Linguistics}.\hskip 1em plus 0.5em minus
  0.4em\relax Association for Computational Linguistics, 2000, pp. 93--100.

\bibitem{williams2007scaling}
J.~D. Williams and S.~Young, ``Scaling pomdps for spoken dialog management,''
  \emph{Audio, Speech, and Language Processing, IEEE Transactions on}, vol.~15,
  no.~7, pp. 2116--2129, 2007.

\bibitem{henderson2008mixture}
J.~Henderson and O.~Lemon, ``Mixture model pomdps for efficient handling of
  uncertainty in dialogue management,'' in \emph{Proceedings of the 46th Annual
  Meeting of the Association for Computational Linguistics on Human Language
  Technologies: Short Papers}.\hskip 1em plus 0.5em minus 0.4em\relax
  Association for Computational Linguistics, 2008, pp. 73--76.

\bibitem{henderson2013deep}
M.~Henderson, B.~Thomson, and S.~Young, ``Deep neural network approach for the
  dialog state tracking challenge,'' in \emph{Proc. SIGDIAL}, 2013.

\bibitem{gasic2014gaussian}
M.~Gasic and S.~Young, ``Gaussian processes for pomdp-based dialogue manager
  optimization,'' \emph{IEEE/ACM Transactions on Audio, Speech and Language
  Processing (TASLP)}, vol.~22, no.~1, pp. 28--40, 2014.

\bibitem{crook2014real}
P.~A. Crook, S.~Keizer, Z.~Wang, W.~Tang, and O.~Lemon, ``Real user evaluation
  of a pomdp spoken dialogue system using automatic belief compression,''
  \emph{Computer Speech \& Language}, vol.~28, no.~4, pp. 873--887, 2014.

\bibitem{polifroni2006learning}
J.~Polifroni and M.~Walker, ``Learning database content for spoken dialogue
  system design,'' in \emph{5th International Conference on Language Resources
  and Evaluation (LREC)}, 2006.

\bibitem{polifroni2008intensional}
J.~Polifroni and M.~A. Walker, ``Intensional summaries as cooperative responses
  in dialogue: Automation and evaluation.'' in \emph{ACL}.\hskip 1em plus 0.5em
  minus 0.4em\relax Citeseer, 2008, pp. 479--487.

\bibitem{pineau2003point}
J.~Pineau, G.~Gordon, S.~Thrun \emph{et~al.}, ``Point-based value iteration: An
  anytime algorithm for pomdps,'' in \emph{IJCAI}, vol.~3, 2003, pp.
  1025--1032.

\bibitem{kemp1997estimating}
T.~Kemp, T.~Schaaf \emph{et~al.}, ``Estimating confidence using word
  lattices.'' in \emph{EuroSpeech}, 1997.

\bibitem{abdou2004beam}
S.~Abdou and M.~S. Scordilis, ``Beam search pruning in speech recognition using
  a posterior probability-based confidence measure,'' \emph{Speech
  Communication}, vol.~42, no.~3, pp. 409--428, 2004.

\bibitem{jiang2005confidence}
H.~Jiang, ``Confidence measures for speech recognition: A survey,''
  \emph{Speech communication}, vol.~45, no.~4, pp. 455--470, 2005.

\bibitem{wang2013simple}
Z.~Wang and O.~Lemon, ``A simple and generic belief tracking mechanism for the
  dialog state tracking challenge: On the believability of observed
  information,'' in \emph{Proceedings of SIGDial}, 2013.

\end{thebibliography}
\end{document}